\documentclass[letterpaper]{article}
\usepackage{dusar}  
\usepackage{times}  
\usepackage{helvet}  
\usepackage{courier}  
\usepackage[hyphens]{url}  
\usepackage{graphicx} 
\usepackage{amsmath} 
\usepackage{natbib}  
\usepackage{caption} 
\usepackage{algorithm}
\usepackage{algorithmic}
\usepackage{multirow} 
\usepackage{booktabs} 
\usepackage{array}
\usepackage{tikz}
\usepackage{pifont}
\usepackage{url}
\usepackage[colorlinks=true, urlcolor=black, citecolor=black, anchorcolor=black, linkcolor=black, pdftex, hidelinks]{hyperref}

\usepackage{newfloat}
\usepackage{listings}
\DeclareCaptionStyle{ruled}{labelfont=normalfont,labelsep=colon,strut=off} 
\floatstyle{ruled}
\newfloat{listing}{tb}{lst}{}
\floatname{listing}{Listing}
\usepackage{xcolor}
\usepackage{seqsplit}
\title{Reflecting with Two Voices: A Co-Adaptive Dual-Strategy Framework for LLM-Based Agent Decision Making}

\author{
    Wentao Zhang\textsuperscript{\rm 1},
    Qunbo Wang\textsuperscript{\rm 2},
    Zhao BoXuan\textsuperscript{\rm 3},
    Tao Zhang\textsuperscript{\rm 1},
    Junsheng Wu\textsuperscript{\rm 1},\\
    HongPin Gan\textsuperscript{\rm 1},
    Ling Dai\textsuperscript{\rm 1},
    ShiZhuang Deng\textsuperscript{\rm 1},
    ShunTong Sun\textsuperscript{\rm 1},
    Yang Liu\textsuperscript{\rm 4},
}

\begin{document}

\maketitle



\footnotetext[1]{%
Wentao Zhang, Tao Zhang, Junsheng Wu, HongPin Gan, Ling Dai, ShiZhuang Deng and ShunTong Sun are with Northwest Polytechnical University, Xi’an, China. 
Email: \href{mailto:wentaozhang@nwpu.edu.cn,taozhang@nwpu.edu.cn,junshengwu@nwpu.edu.cn,hpgan@nwpu.edu.cn,lingdai@nwpu.edu.cn,szdeng@nwpu.edu.cn,stsun@nwpu.edu.cn}{%
  \texttt{\seqsplit{\{wentao\_zhang,dengshizhuang\}@mail.nwpu.edu.cn, \{tao\_zhang, junshengwu, ganhongping\}@nwpu.edu.cn, \{dailing0072,shuntongsun\}@gmail.com}}%
}
}

\footnotetext[2]{%
Qunbo Wang is with Beijing Jiaotong University, Beijing, China. 
Email: \href{mailto:qbwang@bjtu.edu.cn}{%
  \texttt{wangqb6@outlook.com}%
}
}

\footnotetext[3]{%
Yang Liu is with Nanyang Technological University, Singapore. 
Email: \href{mailto:yangliu@ntu.edu.sg}{%
  \texttt{yangliu@ntu.edu.sg}%
}
}
\begin{abstract}
    Large language model (LLM) agents often rely on external demonstrations or retrieval-augmented planning, leading to brittleness, poor generalization, and high computational overhead. Inspired by human problem-solving, we propose DuSAR (Dual-Strategy Agent with Reflecting)—a demonstration-free framework that enables a single frozen LLM to perform co-adaptive reasoning via two complementary strategies: a high-level holistic plan and a context-grounded local policy. These strategies interact through a lightweight reflection mechanism, where the agent continuously assesses progress via a Strategy Fitness Score and dynamically revises its global plan when stuck or refines it upon meaningful advancement, mimicking human metacognitive behavior. On both simulated household (ALFWorld) and real-world web (Mind2Web) environments, DuSAR achieves state-of-the-art performance using only open-source LLMs, substantially outperforming all prior methods without any demonstrations or fine-tuning. Remarkably, it also reduces per-step token consumption by a large margin while maintaining strong task success. Ablation studies confirm the necessity of dual-strategy coordination. Moreover, optional integration of expert demonstrations further boosts performance, highlighting DuSAR's flexibility and compatibility with external knowledge.
\end{abstract}


\section{Introduction}
\begin{figure*}[t]
\centering
\includegraphics[width=0.99\linewidth]{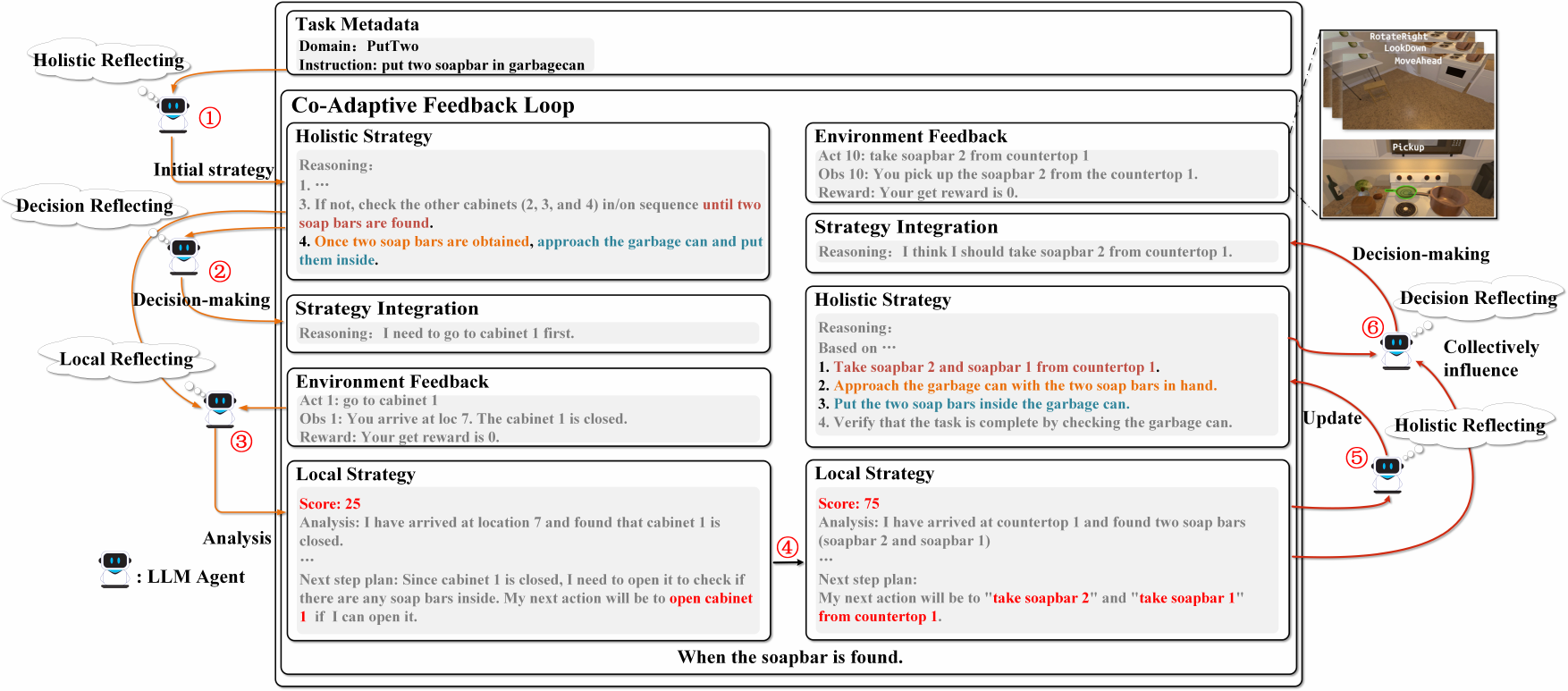}
\caption{The overall architecture of DuSAR, illustrating its iterative dual-strategy reasoning in ALFWorld. Starting from a task instruction, the \textit{Holistic Strategy} generates a high-level plan (\ding{172}), which guides the initial action (\ding{173}). After execution, environmental feedback and the current plan inform the \textit{Local Strategy} to produce context-aware guidance (\ding{174}) and evaluate progress via a \textit{Strategy Fitness Score} (\ding{175}). This evaluation triggers dynamic refinement of the Holistic Strategy (\ding{176}). A \textit{Strategy Integration Module} (SIM) synthesizes both strategies to select the next action (\ding{177}), forming a co-adaptive loop that balances long-term coherence with immediate adaptability. See Fig.~\ref{fig:Framework} for details on the reflecting mechanism.}
\label{fig:Inference-graph}
\end{figure*}

Large language models (LLMs) have shown promise in interactive task planning through reasoning and reflection~\cite{yao2023react, shinn2023reflexion}. Yet, achieving robust performance in long-horizon, partially observable environments remains challenging. Most existing approaches rely on external demonstrations, retrieved exemplars, or auxiliary modules—leading to brittleness, limited generalization, high computational costs, and poor adaptability to novel situations.

A key limitation lies in their dependence on proprietary large language models accessible only via commercial APIs (e.g., GPT-4). While such models often yield strong empirical results, their use introduces significant practical barriers: API calls incur substantial operational costs and raise data privacy concerns, especially in enterprise or regulated settings. Moreover, performance typically degrades markedly when these methods are ported to open-weight, locally deployable LLMs—hindering reproducibility, scalability, and real-world adoption.

Compounding this issue is the heavy reliance on demonstration retrieval, which drastically inflates prompt length—often by several kilotokens per step—severely limiting real-time adaptability, increasing latency, and reducing token efficiency.

This raises a central question: \emph{Can an LLM agent perform adaptive, robust planning without external demonstrations or additional training, using only a frozen base model—and do so with drastically reduced computational overhead?}

We answer affirmatively with \textbf{DuSAR} (\textbf{D}ual-\textbf{S}trategy \textbf{A}gent with \textbf{R}eflecting)—a demonstration-free framework that enables a single frozen LLM to perform co-adaptive reasoning through what we refer to in our title as \textbf{“two voices”}: a \textit{Holistic Strategy} that maintains long-term task structure by decomposing goals into ordered sub-tasks, and a \textit{Local Strategy} that grounds decisions in immediate environmental observations and outputs a \textit{Strategy Fitness Score} quantifying alignment with the current subgoal. These strategies are integrated by a lightweight \textit{Strategy Integration Module} (SIM) within a unified reflective loop. Inspired by human metacognition, the Local Strategy continuously assesses execution progress and signals the Holistic Strategy to revise the global plan upon stagnation or refine it upon meaningful advancement.

Unlike retrieval-based methods that match static exemplars, DuSAR generates and refines structured plan graphs \textit{in situ} through environmental interaction. The Holistic Strategy prevents myopic behavior by preserving global intent, while the Local Strategy enables rapid adaptation to unexpected states or errors. Their co-adaptation yields robust generalization without external supervision and achieves remarkable token efficiency.

We evaluate DuSAR on ALFWorld~\cite{shridhar2021alfworld} and Mind2Web~\cite{deng2023mind2web} using open-source LLMs (7B–70B). Without any in-context examples, DuSAR achieves state-of-the-art results: 37.3\% success on ALFWorld (Llama3.1-70B), outperforming Synapse (13.0\%) and TRAD (9.9\%), and 4.00\% on Mind2Web—more than doubling the best baseline. Ablations confirm the necessity of dual-strategy coordination. Notably, optional integration of expert demonstrations into the Holistic Strategy yields an additional 3–4 percentage point improvement, highlighting DuSAR's flexibility and compatibility with external knowledge. Our contributions are threefold:
\begin{enumerate}
    \item A novel dual-strategy architecture inspired by reflective cognition, enabling a single frozen LLM to perform demonstration-free, co-adaptive planning and execution via a Strategy Integration Module and Strategy Fitness Score evaluation.
    \item A lightweight reflection mechanism that dynamically aligns high-level strategy with low-level observations, achieving strong performance with only 335–564 tokens per step—a 3–9$\times$ efficiency gain over retrieval-based methods.
    \item Extensive empirical validation across benchmarks and model scales, demonstrating generalization, scalability, and seamless compatibility with optional expert demonstrations.
\end{enumerate}

In essence, DuSAR realizes the ‘two voices’ of reflective agency—not as anthropomorphic agents, but as complementary computational strategies within a single LLM.

\begin{figure*}[t]
\centering
\includegraphics[width=0.99\linewidth]{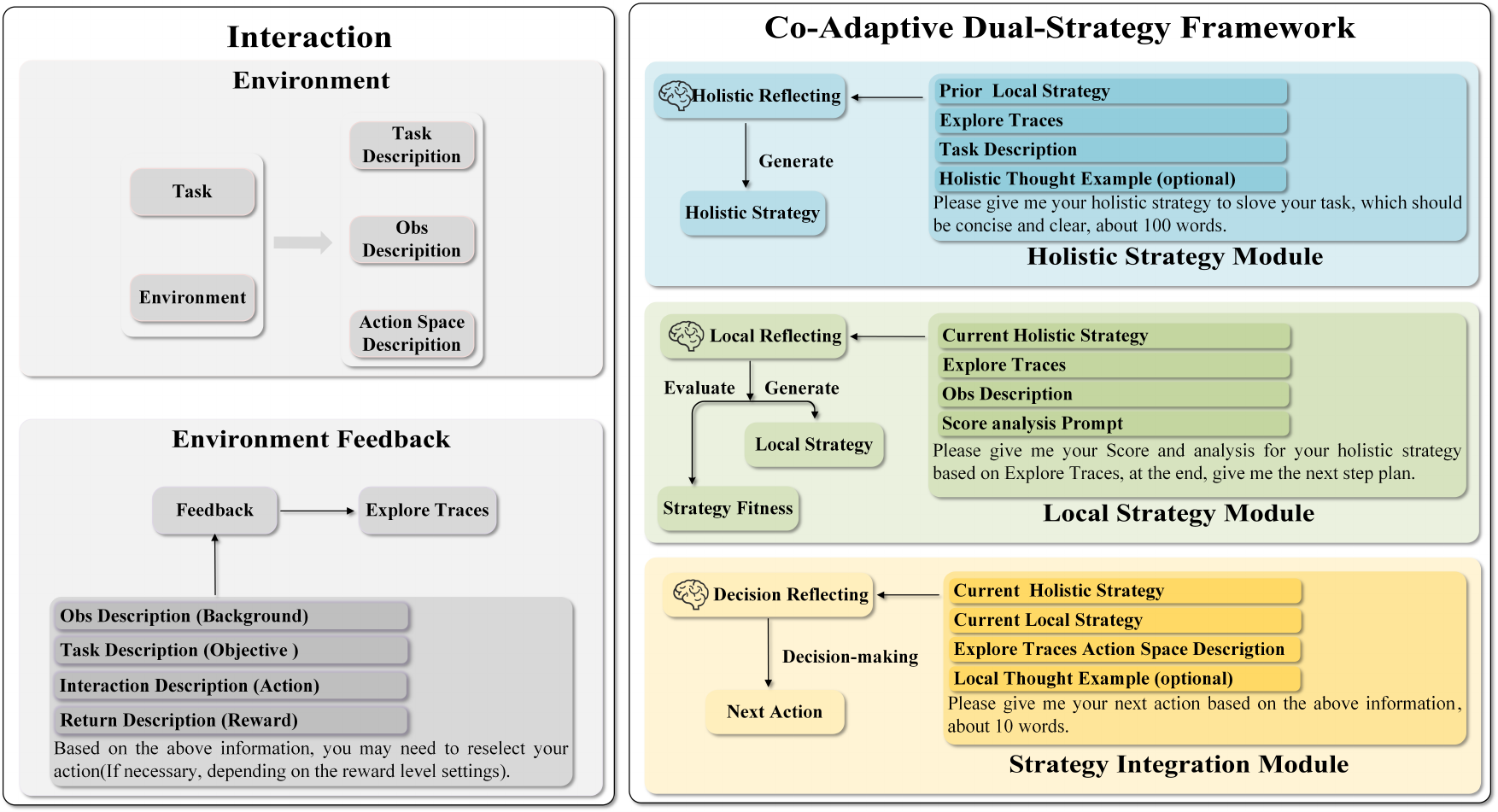}
\caption{DuSAR’s reflecting mechanism. \textit{Holistic Reflecting} formulates a long-term strategy from the task instruction, past exploration traces, and prior local strategies. \textit{Local Reflecting} generates context-aware actions and evaluates the feasibility of the current holistic plan based on environmental observations. \textit{Decision Reflecting} synthesizes both strategies to select the next feasible action. All interactions are logged as structured \textit{Explore Traces} for iterative refinement.}
\label{fig:Framework}
\end{figure*}

\section{Related Work}
\label{sec:related-work}

\subsection{LLM-based Agents and Interactive Planning}
Early LLM agents combined Chain-of-Thought prompting with self-reflection to interleave reasoning and action in text environments~\cite{yao2023react, shinn2023reflexion, yao2023tree}. Recent work extends this to interactive tasks—e.g., web navigation and embodied control—via tool/API integration~\cite{ahn2023do, schick2023toolformer, ICLR2024_28e50ee5}. However, these approaches often falter in long-horizon settings due to challenges in dynamic replanning and partial observability, as evidenced in ALFWorld and WebArena~\cite{shridhar2021alfworld, zhou2024webarena}. Hierarchical methods improve scalability through task decomposition~\cite{wu2023task, petruzzellis2025hierarchical}, but typically depend on manual designs or extensive demonstrations, limiting generalization. Critically, many rely on closed-source models (e.g., GPT-4) and degrade severely on open-source LLMs—ReAct~\cite{yao2023react}, for instance, achieves near-zero success without structured planning, hindering reproducibility and deployment.

\subsection{Generalization and Efficiency Challenges in Agent Design}
Current LLM agents often rely on external support—such as in-context demonstrations~\cite{zhou2024trad, zheng2024synapse} or separate planner/executor modules~\cite{liu2024hierarchical}—which limits their generalization to novel tasks and robustness under distribution shifts~\cite{Yoo_Kim_Woo_2025}. Retrieval-augmented methods like Synapse and TRAD require large demonstration banks, resulting in verbose prompts and high token overhead that hinder scalability and real-time use. Even self-reflection approaches~\cite{madaan2023self} remain dependent on extensive context, incurring significant prompt costs~\cite{wang2025adaptive}. These dependencies underscore a critical gap: the lack of lightweight, zero-shot frameworks that can operate efficiently within a single frozen LLM while maintaining robustness in partially observable environments.

\subsection{Structured and Dual-Strategy Reasoning}
To improve planning robustness, structured approaches have combined global LLM reasoning with local algorithmic modules~\cite{WANG20251020} or hierarchical pipelines~\cite{li2025hierarchical}, inspired by the principle of separating stable planning from adaptive execution. While these methods introduce valuable modularity, they typically require external components, explicit supervision, or additional training—limiting their generality and efficiency. Crucially, they externalize the duality of stable planning and adaptive execution rather than internalizing it within a single model. Drawing inspiration from human metacognitive processes—where strategic planning and tactical execution continuously inform each other through reflective assessment—we abstract the principle of \emph{co-adaptive dual-strategy reasoning} into a purely prompt-based framework, internalized within a single frozen LLM. DuSAR realizes this by instantiating two co-adaptive reasoning strategies (Holistic and Local Strategies) within a frozen LLM, enabling dynamic strategy refinement through Strategy Fitness Scores without external modules or training.

\section{Methodology}
\label{sec:method}

We propose the \emph{\textbf{Du}al-\textbf{S}trategy \textbf{A}gent with \textbf{R}eflecting} (DuSAR), a demonstration-free, zero-shot framework that internalizes a dual-strategy reasoning paradigm within a frozen large language model (LLM) to enable robust, long-horizon planning in interactive environments. As illustrated in Fig.~\ref{fig:Framework}, DuSAR comprises three tightly coupled modules—\textit{Holistic Reflecting}, \textit{Local Reflecting}, and \textit{Decision Reflecting}—which collectively implement a closed-loop, co-adaptive decision-making process without requiring expert demonstrations, in-context examples, or parameter updates. In DuSAR, \textit{Reflecting} denotes the internal cognitive process that generates insights from experience, while \textit{Strategy} refers to the structured plan or action directive it produces.

Inspired by human metacognition and dual-network architectures in reinforcement learning, DuSAR decouples strategic planning from tactical execution through two complementary strategies:
\begin{itemize}
    \item A \textit{Holistic Strategy} $H_t$, which maintains a high-level, temporally extended plan for task decomposition and long-term coherence;
    \item A \textit{Local Strategy} $L_t$, which generates context-sensitive actions and evaluates immediate progress.
\end{itemize}
The Holistic Strategy acts as a stable reference that guides and constrains the Local Strategy, preventing myopic behavior and unproductive loops. In turn, the Local Strategy provides real-time feedback via a \textit{Strategy Fitness Score}, which triggers dynamic updates to the Holistic Strategy upon stagnation or milestone achievement. This bidirectional interaction forms a reflective, self-correcting loop that balances global consistency with local adaptability.

\subsection{Formal Specification}
\label{subsec:formal}

Let $I$ denote the initial task instruction, and let $E_{<t} = \{(o_i, a_i, r_i, l_i, s_i)\}_{i=1}^{t-1}$ represent the \textit{Explore Trace} up to step $t-1$, where $o_i$, $a_i$, $r_i$, $l_i$, and $s_i$ are the observation, action, reward, local log, and Strategy Fitness Score at step $i$, respectively.

\textbf{Holistic Strategy Update:}
For $t = 1$, we set $H_1 = H_0$, where $H_0$ is initialized from the task instruction $I$. For $t \geq 2$, the Holistic Strategy $H_t$ is updated at the beginning of step $t$ based on the previous step's fitness score $s_{t-1}$:
\begin{equation}
    H_t = 
    \begin{cases}
        H_{\text{Ref}}(I, E_{<t}, H_{t-1}) & \text{if } s_{t-1} = 0 \text{ or } 50 \leq s_{t-1} \leq 99 \\
        H_{t-1} & \text{if } 1 \leq s_{t-1} \leq 49 \\
        \text{Terminate} & \text{if } s_{t-1} = 100
    \end{cases}
    \label{eq:holistic_update}
\end{equation}
Here, $H_0$ is initialized from $I$, and $H_{\text{Ref}}$ denotes the \textit{Holistic Reflecting} function. The update logic encodes four semantic states based on $s_{t-1}$: error recovery ($s_{t-1}=0$), plan maintenance ($1 \leq s_{t-1} \leq 49$), sub-goal advancement ($50 \leq s_{t-1} \leq 99$), and task completion ($s_{t-1}=100$). By updating $H_t$ at the start of step $t$, the updated strategy can guide decision-making throughout the current step.

\textbf{Local Strategy Generation and Reasoning Log:}
At each step $t$, the Local Strategy and its associated reasoning log are generated based on the current observation:
\begin{equation}
    (L_t, l_t) = L_{\text{Ref}}(o_t, H_t, E_{<t}),
    \label{eq:local}
\end{equation}
where $L_{\text{Ref}}$ denotes \textit{Local Reflecting}, and $l_t$ is the reasoning log that records the local reasoning process.

\textbf{Action Selection:}
The action is selected by integrating both strategies:
\begin{equation}
    a_t = D_{\text{Ref}}(H_t, L_t, o_t, \mathcal{A}_t),
    \label{eq:action}
\end{equation}
where $D_{\text{Ref}}$ denotes \textit{Decision Reflecting}. The reasoning log $l_t$, generated by Local Reflecting, is recorded in the trace but not used in action selection.

\textbf{Progress Assessment:}
After executing $a_t$ and receiving reward $r_t$, the Strategy Fitness Score is computed:
\begin{equation}
    s_t = S_{\text{Ana}}(o_t, a_t, r_t, E_{<t}) \in [0, 100],
    \label{eq:score}
\end{equation}
where $S_{\text{Ana}}$ denotes the progress assessment function, elicited from the LLM using a fixed, handcrafted prompt (see Appendix for details). The Strategy Fitness Score is interpreted as follows:
\begin{itemize}
    \item $s_t = 0$: no meaningful progress;
    \item $1 \leq s_t \leq 49$: ongoing exploration within current sub-goal;
    \item $50 \leq s_t \leq 99$: significant milestone achieved;
    \item $s_t = 100$: full task completion.
\end{itemize}

\textbf{Trace Update:}
The Explore Trace is updated with the complete step information:
\begin{equation}
    E_t = E_{<t} \cup \{(o_t, a_t, r_t, l_t, s_t)\},
    \label{eq:trace}
\end{equation}
where all components are defined as above. Integration prioritizes the Holistic Strategy when aligned with local context; otherwise, it defers to the Local Strategy for grounded adaptation.

\begin{algorithm}[t]
\caption{DuSAR Co-Adaptive Decision Loop}
\label{alg:dusar}
\begin{algorithmic}[1]
\REQUIRE Task instruction $I$, initial observation $o_0$
\STATE $H_0 \leftarrow \text{HolisticReflect}(I, \emptyset, \emptyset)$, $E \leftarrow \emptyset$
\FOR{step $t = 1, 2, \ldots$ until task completion or max steps}
    \STATE Observe $o_t$ from environment
    \IF{$t = 1$}
        \STATE Set $H_t \leftarrow H_0$ \COMMENT{For $t=1$, use initialized strategy}
    \ELSIF{($s_{t-1} = 0$) \OR ($50 \leq s_{t-1} \leq 99$)}
        \STATE Update $H_{t} \leftarrow \text{HolisticReflect}(I, E, H_{t-1})$
    \ELSIF{$s_{t-1} = 100$}
        \STATE Task completed; terminate
    \ELSE
        \STATE Maintain $H_t \leftarrow H_{t-1}$
    \ENDIF
    \STATE Generate $(L_t, l_t) \leftarrow \text{LocalReflect}(o_t, H_t, E)$ \COMMENT{$l_t$ is the reasoning log}
    \STATE Execute $a_t \leftarrow \text{DecisionReflect}(H_t, L_t, o_t, \mathcal{A}_t)$
    \STATE Receive reward $r_t$ and update observation
    \STATE Compute $s_t \leftarrow \text{ScoreAnalysis}(o_t, a_t, r_t, E)$ \COMMENT{$s_t \in [0, 100]$}
    \STATE Append to trace: $E \leftarrow E \cup \{(o_t, a_t, r_t, l_t, s_t)\}$
\ENDFOR
\end{algorithmic}
\end{algorithm}

\begin{table*}[htbp]
\centering
\small 
\renewcommand{\arraystretch}{1.0} 
\setlength{\tabcolsep}{6.0pt} 

    \caption{Success rates across six ALFWorld task types. DuSAR achieves significantly higher success rates than all baselines overall, with a particularly pronounced improvement on the \textit{PutTwo} sub-task—where baseline methods achieve near-zero success—demonstrating its superior capability in generating effective, multi-step action sequences for complex object manipulation. A success rate of 0.0 indicates failure to complete the task within the step limit.}
\label{tab:comparison_new}
\begin{tabular}{l l cccccc c} 
\toprule
Model & Method & Put & Examine & Clean & Heat & Cool & PutTwo & All SR \\
\midrule
\multirow{3}{*}{\shortstack{Llama2-7B}} 
& Synapse & 0.0 & 0.0 & 0.0 & 0.0 & 0.0 & 0.0 & 0.0 \\
& TRAD & 0.0 & 0.0 & 0.0 & 0.0 & 0.0 & 0.0 & 0.0 \\
& \bfseries DuSAR (Ours) & \bfseries 4.2 &  0.0 &  0.0 &  0.0 &  0.0 & 0.0 & \bfseries 0.75 \\ 

\midrule
\multirow{3}{*}{\shortstack{Llama3.1-8B}}
    & Synapse & 4.2 & 16.7 & 0.0 & 0.0 & 0.0 & 0.0 & 3.5 \\
& TRAD & 0.0 & 0.0 & 0.0 & 0.0 & 0.0 & 0.0 & 0.0 \\
    & \bfseries DuSAR (Ours) & \bfseries 37.5 & \bfseries 22.2 & 0.0 & \bfseries 4.3 &  0.0 & \bfseries 5.9 & \bfseries 11.7 \\

\midrule
\multirow{3}{*}{\shortstack{Gemma3-12B}}
    & Synapse & 12.5 & 5.5 & \bfseries 16.1 & \bfseries 47.8 & \bfseries 4.8 & 0.0 & 14.5 \\
& TRAD & 8.3 & 11.1 & 0.0 & 34.8 & 0.0 & 0.0 & 9.0 \\
& \bfseries DuSAR (Ours) & \bfseries 45.8 & \bfseries 27.8 & 12.9 & 13.0 & \bfseries 4.8 & \bfseries 11.8 & \bfseries 19.4 \\

\midrule
\multirow{3}{*}{\shortstack{Llama3.1-70B}}
    & Synapse & 12.5 & 16.7 & 12.9 & \bfseries 26.1 & 9.5 & 0.0 & 13.0 \\
    & TRAD & 12.5 & 11.1 & 3.2 & 8.7 & \bfseries 23.8 & 0.0 & 9.9 \\
    & \bfseries DuSAR (Ours) & \bfseries 75.0 & \bfseries 61.1 & \bfseries 19.4 & 13.0 & 14.3 & \bfseries 52.9 & \bfseries 37.3 \\
\bottomrule
\end{tabular}
\end{table*}

\subsection{Holistic Strategy Module}
The Holistic Strategy addresses the challenge of long-horizon coherence in partially observable environments. By decomposing the task into ordered sub-goals (e.g., "Find two soapbars" for "put two soapbars in garbagecan", as in Fig.~\ref{fig:Inference-graph}), it provides a stable scaffold that prevents the agent from drifting into unproductive loops—a failure mode that arises when agents rely solely on local, reactive planning without global structure. Updates are triggered only upon meaningful feedback ($s_t=0$ or $50\leq s_t\leq99$), ensuring plan stability during active exploration.

\subsection{Local Strategy Module}
While the Holistic Strategy ensures global consistency, the Local Strategy enables rapid adaptation to environmental dynamics. Critically, \textit{Local Reflecting} jointly generates actions and evaluates progress via the Strategy Fitness Score $s_t$, which quantifies advancement without external supervision. This self-assessment capability allows DuSAR to detect stagnation (e.g., repeated failed interactions) and trigger replanning—eliminating the need for handcrafted reward shaping or demonstration-based feedback.

\subsection{Strategy Integration Module}
Naive fusion of global and local signals (e.g., prompt concatenation) often leads to conflict or dilution of intent, as the two strategies may provide contradictory guidance. Our priority-based integration in \textit{Decision Reflecting} resolves this by adhering to the Holistic Strategy when aligned, but yielding to local grounding under uncertainty—striking a balance essential for robustness in dynamic environments.

\subsection{Environment Feedback and Explore Traces}
\label{subsec:traces}
All modules share a common memory—the \textit{Explore Trace} $E_{<t}$—which records $(o_t, a_t, r_t, l_t, s_t)$ tuples. A sliding window ($K=10$) retains only recent steps, ensuring computational efficiency while preserving sufficient history for coherent reasoning.

\section{Experiments}
\label{sec:exp}

\subsection{Experimental Setup}
To assess whether an LLM agent can achieve robust, generalizable planning without external demonstrations, additional training, or proprietary APIs—and at lower computational cost—we compare DuSAR against two state-of-the-art demonstration-driven agents: \textbf{Synapse}~\cite{zheng2024synapse} and \textbf{TRAD}~\cite{zhou2024trad}. Synapse uses single-stage trajectory-level retrieval based on task similarity, while TRAD employs a two-stage approach: first retrieving task-level exemplars, then refining based on intermediate reasoning (``thought'') similarity. Both rely on large external demonstration banks and are typically evaluated with proprietary models like GPT-4. In contrast, DuSAR requires no demonstrations and is designed for open-source LLMs.

We also include \textbf{ReAct}~\cite{yao2023react} as a baseline that uses fixed in-context examples (two per task type). However, ReAct achieves negligible performance on open-source models (at most 1.49\% on ALFWorld with Llama3.1-70B, and 0.13\% on Mind2Web), leading us to omit its ALFWorld results from Table~\ref{tab:comparison_new} due to effectively zero success across most model scales.

We evaluate all methods on two challenging benchmarks. \textbf{ALFWorld}~\cite{shridhar2021alfworld} is a text-based household simulation requiring long-horizon planning, object affordance reasoning, and exploration under partial observability. Following standard protocol~\cite{shridhar2021alfworld}, we test on 134 out-of-distribution tasks across six sub-task categories—\emph{Put}, \emph{Examine}, \emph{Clean}, \emph{Heat}, \emph{Cool}, and \emph{PutTwo}—and report overall success rate (All SR).

\textbf{Mind2Web}~\cite{deng2023mind2web} requires agents to complete real-world web tasks (e.g., booking a hotel) by interacting with raw HTML DOMs. We evaluate under three cross-generalization settings: \textit{Cross-Task}, \textit{Cross-Website}, and \textit{Cross-Domain}, along with their macro-average (\textit{All}). Metrics include Element Accuracy (Ele. Acc), Step-level Success Rate (Step SR), and Task Success Rate (Task SR). For fair comparison with prior work~\cite{zhou2024trad}, we summarize observations into 5 salient web elements using the pre-trained ranker from Deng et al.~\cite{deng2023mind2web}.

All methods are evaluated across four open-source LLMs: \texttt{Llama2-7B}, \texttt{Llama3.1-8B}, \texttt{Gemma3-12B}, and \texttt{Llama3.1-70B}, with decoding hyperparameters set to temperature=0, top-$p$=0.8, presence penalty=0.1, and frequency penalty=0.1.

\begin{table*}[htbp]
\centering
    \caption{Results(\%) of all methods on Mind2Web benchmark. DuSAR achieves state-of-the-art performance across diverse LLMs (7B--70B), outperforming Synapse, TRAD, and ReAct in task success rate and element accuracy under cross-domain generalization settings, with consistent gains across most model scales and metrics.}
\small
\setlength{\tabcolsep}{3.5pt}
\begin{tabular}{l l *{12}{c}}
\toprule
\multirow{2}{*}{Model} & \multirow{2}{*}{Method} & \multicolumn{3}{c}{Cross-Task} & \multicolumn{3}{c}{Cross-Website} & \multicolumn{3}{c}{Cross-Domain} & \multicolumn{3}{c}{All} \\
\cmidrule(lr){3-5} \cmidrule(lr){6-8} \cmidrule(lr){9-11} \cmidrule(l){12-14}
& & Ele. Acc & Step SR & SR & Ele. Acc & Step SR & SR & Ele. Acc & Step SR & SR & Ele. Acc & Step SR & SR \\
\midrule

\multirow{4}{*}{Llama2-7B}
    & ReAct & 7.2 &  6.0 & 0.00 & 8.1 & 6.6 & 0.00 & 6.6 &  5.2 & 0.11 & 7.3 & 5.9 & 0.04 \\
    & Synapse & 12.3 & 9.6 & 0.00 & 12.0 & \bfseries 9.5 & 0.00 & \bfseries 13.0 & 8.6 & \bfseries 0.22 & 12.4 & 9.2 & 0.07 \\
    & TRAD & 11.4 & 9.1 & 0.00 & 9.9 & 7.6 & 0.00 & 10.6 & 7.4 & 0.00 & 10.6 & 8.0 & 0.00 \\
    & \bfseries DuSAR (Ours) & \bfseries 15.1 & \bfseries 12.6 & 0.00 & \bfseries 12.8 & 6.6 & 0.00 & 12.5 & \bfseries 9.8 & 0.11 & \bfseries 13.5 & \bfseries 9.7 & 0.04 \\

\midrule
    
\multirow{4}{*}{Llama3.1-8B}
    & ReAct & 7.1 & 3.5 & 0.00 & 6.2 & 3.1 & 0.00 & 5.7 & 3.1 & 0.00 & 6.3 & 3.2 & 0.00 \\
    & Synapse & 22.7 & 9.8 & 0.00 & 18.2 & 6.7 & 0.00 & 21.1 & 9.6 & 0.33 & 20.7 & 8.7 & 0.11 \\
    & TRAD & 17.4 & 13.6 & 0.0 & 15.1 & 9.7 & 0.00 & 15.4 & 10.1 & 0.05 & 16.0 & 11.1 & 0.02 \\
    & \bfseries DuSAR (Ours) & \bfseries 48.2 & \bfseries 34.3 & \bfseries 1.98 & \bfseries 37.8 & \bfseries 25.4 & \bfseries 0.56 & \bfseries 35.1 & \bfseries 31.0 & \bfseries 3.00 & \bfseries 43.9 & \bfseries 31.6 & \bfseries 1.76 \\

\midrule
    
\multirow{4}{*}{Gemma3-12B}
    & ReAct & 23.4 & 17.3 & 0.40 & 20.1 & 14.0 & 0.56 & 21.8 & 15.4 & 0.33 & 21.8 & 15.6 & 0.43 \\
    & Synapse & 31.0 & 26.1 & 1.19 & 27.3 & 20.7 & 1.13 & 30.5 & 22.5 & 0.62 & 29.6 & 23.1 & 0.98 \\
    & TRAD & 34.8 & 31.1 & 1.98 & 28.2 & 23.0 & 0.00 & 28.5 & 24.6 & 0.88 & 30.5 & 26.2 & 0.95 \\
    & \bfseries DuSAR (Ours) & \bfseries 53.2 & \bfseries 45.2 & \bfseries 5.21 & \bfseries 37.9 & \bfseries 29.6 & \bfseries 1.51 & \bfseries 47.4 & \bfseries 39.4 & \bfseries 4.45 & \bfseries 46.1 & \bfseries 38.1 & \bfseries 3.72 \\

\midrule
    
\multirow{4}{*}{Llama3.1-70B}
    & ReAct & 15.2 & 12.0 & 0.00 & 10.7 & 8.5 & 0.00 & 10.8 & 8.9 & 0.38 & 12.2 & 9.8 & 0.13 \\
    & Synapse & 35.0 & 22.0 & 1.19 & 29.4 & 19.0 & 0.00 & 29.5 & 17.3 & 0.44 & 31.3 & 19.4 & 0.54 \\
    & TRAD & 44.2 & 37.6 & 2.38 & 36.0 & 29.0 & 0.85 & 38.8 & 33.5 & 1.94 & 39.6 & 33.4 & 1.75 \\
    & \bfseries DuSAR (Ours) & \bfseries 54.9 & \bfseries 45.1 & \bfseries 5.28 & \bfseries 44.6 & \bfseries 32.3 & \bfseries 3.11 & \bfseries 45.8 & \bfseries 36.8 & \bfseries 3.61 & \bfseries 48.4 & \bfseries 38.1 & \bfseries 4.00 \\

\bottomrule
\end{tabular}
\begin{minipage}{\linewidth}
\footnotesize
$^*$SR denotes Task Success Rate (Task SR), measuring complete task completion, distinct from Step SR which measures per-step action correctness.
\end{minipage}
\label{tab:comparison_Min2Web}
\end{table*}

\subsection{Evaluation on ALFWorld}
As shown in Table~\ref{tab:comparison_new}, DuSAR establishes a new state of the art across all model scales. With Llama3.1-70B, it achieves 37.3\% overall success rate—more than 2.9$\times$ higher than Synapse (13.0\%) and TRAD (9.9\%). Crucially, DuSAR enables non-trivial performance on small models: 11.7\% with Llama3.1-8B and 19.4\% with Gemma3-12B, whereas Synapse and TRAD achieve near-zero success ($<$0.1\%) on 7B–8B scales. The performance gap stems from fundamental limitations of retrieval-based approaches in interactive, dynamic environments.

\paragraph{Failure Modes of Retrieval-Based Agents on Small Models}
In ALFWorld’s interactive, continuously evolving environment, retrieval-based agents achieve only 0–3.5\% success on 7B–8B models despite augmentation. This stems from: (1) \textit{instruction–exemplar conflation}, where small models treat retrieved trajectories as literal instructions; (2) \textit{trajectory mismatch under dynamics}, as fixed demonstrations rarely align with the agent’s actual state trajectory; and (3) \textit{excessive token use} (1.5k–3.6k/step), which crowds out real-time reasoning and error correction in limited contexts.

\paragraph{Demonstration-Free Structured Planning}
DuSAR overcomes these limitations by performing planning without external demonstrations. It employs an internal dual-strategy architecture: a Holistic Strategy that decomposes tasks into abstract sub-goals, and a Local Strategy that grounds each action in current observations and refines execution via feedback. This lightweight yet structured reasoning transforms raw LLM outputs into adaptive, self-contained plans—avoiding brittle replay of exemplar behaviors while operating efficiently within tight token budgets (335–564 tokens/step).

\paragraph{Structured Reasoning as a Long-Horizon Planning Enabler}
DuSAR shows that effective long-horizon planning stems not from memorizing full trajectories, but from a dynamic, hierarchical plan structure. The Holistic Strategy maintains high-level intent across steps, while the Local Strategy enables local adaptation and recovery—yielding a closed-loop planner robust to partial observability and execution errors. In contrast, retrieval-based methods treat planning as pattern matching and fail when environments diverge from stored examples. This explains DuSAR’s strong scaling: even small models reason coherently once freed from the noise and rigidity of external demonstrations.

\paragraph{Robustness on Long-Horizon Tasks}
The \emph{PutTwo} task requires six sequential actions (find $\rightarrow$ pick $\rightarrow$ go $\rightarrow$ place, repeated twice)—a challenge for trajectory-matching approaches. Synapse and TRAD achieve 0\% success across all model scales, as they attempt to retrieve or mimic full six-step sequences, which rarely generalize. DuSAR, by contrast, decomposes the task into reusable sub-goals and adapts step-by-step, achieving 52.9\% success with Llama3.1-70B, 11.8\% with Gemma3-12B, and 5.9\% with Llama3.1-8B.

\paragraph{Limitations and Failure Cases}
DuSAR’s performance depends on task clarity and observation quality. It underperforms when (1) observations are ambiguous (e.g., occluded objects), (2) tasks require highly specialized knowledge best captured by curated demonstrations (e.g., ALFWorld’s \emph{Heat} subtask), or (3) goals are too vague for meaningful decomposition (e.g., “find something interesting”). In such cases, the reasoning scaffold remains functional but lacks sufficient signal to construct a valid plan—highlighting that structured reasoning complements, but does not replace, well-specified tasks. 

\begin{table*}[htbp]
\centering
\caption{Average token consumption per step on Mind2Web. DuSAR drastically reduces Prompt Tokens (To\_Pro) by 3--9$\times$ compared to retrieval-based methods (335--564 vs. 1.5k--3.7k tokens/step) while maintaining meaningful Completion Tokens (To\_Com), indicating effective action generation.}
\small
\setlength{\tabcolsep}{6pt}
\begin{tabular}{l l *{8}{c}}
\toprule
\multirow{2}{*}{Model} & \multirow{2}{*}{Method} & \multicolumn{2}{c}{Cross-Task} & \multicolumn{2}{c}{Cross-Website} & \multicolumn{2}{c}{Cross-Domain} & \multicolumn{2}{c}{All} \\
\cmidrule(lr){3-4} \cmidrule(lr){5-6} \cmidrule(lr){7-8} \cmidrule(l){9-10}
& & To\_Pro & To\_Com & To\_Pro & To\_Com & To\_Pro & To\_Com & To\_Pro & To\_Com \\
\midrule

\multirow{4}{*}{Llama2-7B}
& ReAct & 1914.8 & 111.1 & 1901.0 & 81.7 & 1853.8 & 100.0 & 1889.9 & 97.6 \\
& Synapse & 2124.5 & 36.2 & 2078.9 & 41.3 & 1998.2 & 33.3 & 2067.2 & 36.9 \\
& TRAD & 3589.9 & 109.3 & 3611.2 & 287.3 & 3594.8 & 91.6 & 3598.6 & 162.7 \\
& \bfseries DuSAR (Ours) & \bfseries 539.5 & \bfseries 80.5 & \bfseries 551.0 & \bfseries 79.3 & \bfseries 563.9 & \bfseries 80.9 & \bfseries 551.5 & \bfseries 80.2 \\

\midrule
\multirow{4}{*}{Llama3.1-8B}
& ReAct & 1728.0 & 626.3 & 1708.4 & 762.6 & 1625.3 & 514.0 & 1687.2 & 634.3 \\
& Synapse & 1909.2 & 20.5 & 2059.8 & 56.6 & 1961.8 & 68.4 & 1976.9 & 48.5 \\
& TRAD & 3302.0 & 657.8 & 3323.3 & 548.2 & 3258.4 & 442.2 & 3294.6 & 549.4 \\
& \bfseries DuSAR (Ours) & \bfseries 381.9 & \bfseries 16.2 & \bfseries 411.0 & \bfseries 11.7 & \bfseries 334.9 & \bfseries 9.7 & \bfseries 375.9 & \bfseries 12.5 \\

\midrule
\multirow{4}{*}{Gemma3-12B}
& ReAct & 1571.2 & 12.5 & 1557.9 & 13.9 & 1501.2 & 11.8 & 1543.4 & 12.7 \\
& Synapse & 1777.2 & 18.5 & 1854.3 & 21.1 & 1663.0 & 19.7 & 1764.8 & 19.8 \\
& TRAD & 3369.2 & 19.4 & 3407.9 & 19.6 & 3333.2 & 19.3 & 3370.1 & 19.4 \\
& \bfseries DuSAR (Ours) & \bfseries 393.4 & \bfseries 11.3 & \bfseries 408.3 & \bfseries 18.5 & \bfseries 375.8 & \bfseries 10.4 & \bfseries 392.5 & \bfseries 13.4 \\

\midrule
\multirow{4}{*}{Llama3.1-70B}
& ReAct & 1482.7 & 437.1 & 1477.3 & 315.6 & 1422.8 & 341.4 & 1460.9 & 364.7 \\
& Synapse & 1545.8 & 18.5 & 2059.8 & 57.8 & 1961.8 & 68.5 & 1855.8 & 48.3 \\
& TRAD & 3244.1 & 30.0 & 3284.8 & 31.1 & 3324.6 & 23.8 & 3284.5 & 28.3 \\
& \bfseries DuSAR (Ours) & \bfseries 378.2 & \bfseries 10.7 & \bfseries 364.1 & \bfseries 11.8 & \bfseries 348.4 & \bfseries 10.3 & \bfseries 363.6 & \bfseries 10.9 \\

\bottomrule
\end{tabular}
\label{tab:token_efficiency_mind2web}
\end{table*}

\subsection{Evaluation on Mind2Web}
As shown in Tables~\ref{tab:comparison_Min2Web} and~\ref{tab:token_efficiency_mind2web}, DuSAR achieves superior performance across all model scales and generalization settings. With Llama3.1-70B, it attains 4.00\% Task Success Rate (SR). Note that Mind2Web Task SR is a rigorous metric; for comparison, Synapse and ReAct achieve effectively 0\% success on these open-weight scales. DuSAR's result is more than 2$\times$ higher than TRAD (1.75\%) and over 7$\times$ Synapse (0.54\%). Crucially, DuSAR enables non-trivial cross-domain generalization even on small models: 3.00\% (8B) and 4.45\% (12B), while baselines achieve at most 0.33\% (Synapse) and often as low as 0.00\%.

\paragraph{Generalization Challenges for Retrieval-Based Agents}
Retrieval-based methods degrade severely in cross-domain settings due to the abrupt, non-stationary nature of web interfaces. On Llama2-7B and Llama3.1-8B, TRAD achieves only 0.00\% and 0.05\% Task SR—far below DuSAR’s 0.11\% and 3.00\%. This stems from: (1) \textit{domain mismatch}, where training-domain exemplars embed irrelevant UI patterns; (2) \textit{exemplar-task conflation}, as interleaved demonstrations confuse small models about applicable behaviors; and (3) \textit{limited abstraction}, causing models to memorize surface UI structures instead of learning transferable action logic, leading to catastrophic failure under distribution shift.

\paragraph{Domain-Agnostic Planning via Internal Abstraction}
DuSAR addresses these challenges by eschewing external exemplars and instead constructing domain-agnostic plans through internal reasoning. Its Holistic Strategy compresses noisy, domain-specific HTML into abstract, goal-oriented sub-tasks, while the Local Strategy grounds each action in real-time observations. This dual-strategy architecture enables consistent interpretation of diverse web interfaces without relying on retrieved patterns, allowing DuSAR to generalize effectively across unseen domains—even on small models.

\paragraph{Structured Reasoning as a Generalization Scaffold}
DuSAR shows that robust cross-domain generalization arises not from more demonstrations, but from an internal reasoning scaffold that decouples task structure from surface UI patterns. By decomposing tasks into abstract sub-goals (Holistic Strategy) and grounding actions via real-time feedback (Local Strategy), it constructs domain-invariant plans—enabling coherent execution across unseen interfaces like e-commerce, healthcare, and government sites. In contrast, retrieval-based methods conflate task logic with exemplar-specific layouts, leading to brittle failures when UI cues shift. This explains DuSAR’s consistent performance on small models and under distribution shift, while demonstration-dependent approaches collapse even with larger capacity.

\paragraph{Robust Execution in Dynamic Web Environments}
Mind2Web's dynamic HTML—where page structure shifts abruptly between steps—challenges plan coherence. DuSAR's closed-loop design adapts by maintaining high-level intent (Holistic Strategy) and correcting actions using current DOM states (Local Strategy). Under cross-domain evaluation, it achieves 3.00\% Task SR on Llama3.1-8B and 4.45\% on Gemma3-12B—over 2$\times$ TRAD's performance. Notably, TRAD reaches only 2.1\% average Task SR even with GPT-3.5-Turbo~\cite{zhou2024trad}, confirming that the bottleneck lies in reasoning architecture, not model capacity.

\paragraph{Token Efficiency Enables Practical Deployment}
By eliminating demonstration retrieval, DuSAR drastically reduces prompt length. As Table~\ref{tab:token_efficiency_mind2web} shows, it uses only 335–564 tokens per step (To\_Pro), which is 3–9$\times$ fewer than Synapse/TRAD (1.5k–3.7k) and ReAct (1.4k–2.2k). Remarkably, this efficiency comes with no performance trade-off: DuSAR achieves 4.00\% Task SR on Llama3.1-70B (a strong result on the rigorous Mind2Web benchmark) versus 0.13\% for ReAct. This demonstrates that compact, structured reasoning outperforms verbose, exemplar-heavy prompting. 

\begin{figure}[htbp]
\centering
\includegraphics[width=0.48\textwidth]{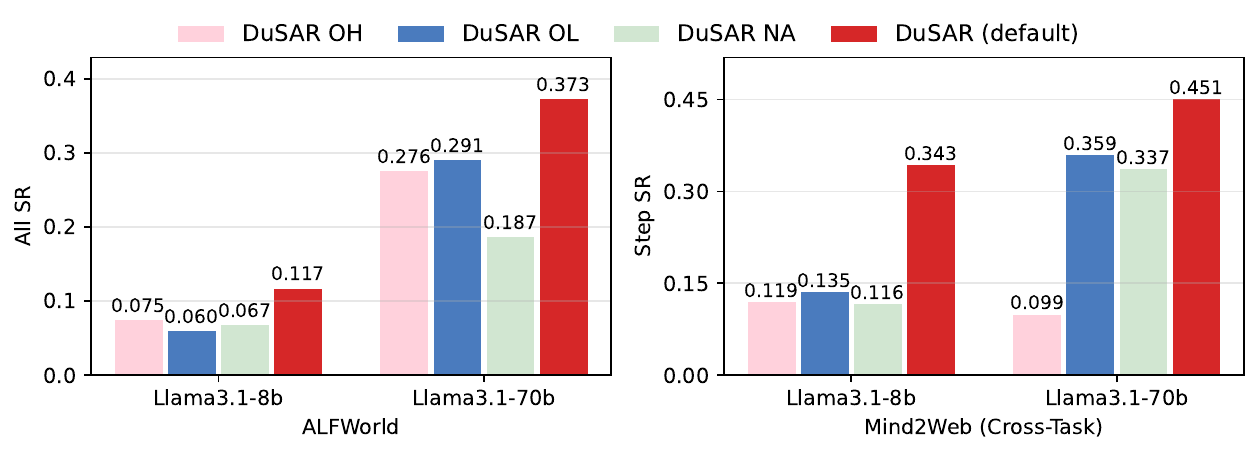}
\caption{Module ablation: Holistic-only (DuSAR OH), Local-only (DuSAR OL), and naive concatenation (DuSAR NA) on ALFWorld (All SR) and Mind2Web Cross-Task (Step SR).}
\label{fig:global_local_ablation}
\end{figure}

\subsection{Ablation Studies}
We conduct ablations using Llama3.1-8B and Llama3.1-70B to (1) validate the necessity of DuSAR’s co-adaptive dual-strategy, and (2) assess whether external demonstrations can improve upon its zero-shot design.

\paragraph{Necessity of Co-Adaptive Dual-Strategy Architecture}
We evaluate three module ablation variants: \textbf{DuSAR OH} (Holistic-only), \textbf{DuSAR OL} (Local-only), and \textbf{DuSAR NA} (naive concatenation without co-adaptation). 

On ALFWorld (All SR), both OH and OL improve with model scale (0.075→0.28 for OH; 0.060→0.29 for OL from 8B to 70B), indicating that larger models can partially compensate for the absence of one reasoning pathway. However, \textbf{NA}, which combines both strategies without dynamic alignment, peaks at only 0.19 SR, far below full DuSAR (0.373 SR). This demonstrates that co-adaptive integration is more critical than mere component presence.

On Mind2Web (Step SR), the roles diverge: the Local Strategy (\textbf{OL}) scales effectively (0.14→0.36 Step SR), while the Holistic Strategy alone (\textbf{OH}) shows diminishing returns with scale (0.12→0.10 Step SR), suggesting that abstract planning requires local grounding to remain effective in complex, non-stationary web interfaces. Although \textbf{NA} improves with capacity (0.12→0.34 Step SR), it still lags behind the default co-adaptive design (0.451 Step SR). Crucially, on Llama3.1-8B, all ablated variants achieve at most 0.14 Step SR---less than half of DuSAR's 0.343---confirming that robust performance stems from co-adaptive reasoning, not model scale alone.

\paragraph{Compatibility with External Demonstrations}
We evaluate DuSAR's compatibility with external demonstrations via three variants: \textbf{DuSAR HT} augments the Holistic Strategy with expert sub-goal traces; \textbf{DuSAR LT} injects expert action traces into the Local Strategy; \textbf{DuSAR BT} combines both—representing a "best-of-TRAD" setup.

\begin{figure}[htbp]
\centering
\includegraphics[width=0.48\textwidth]{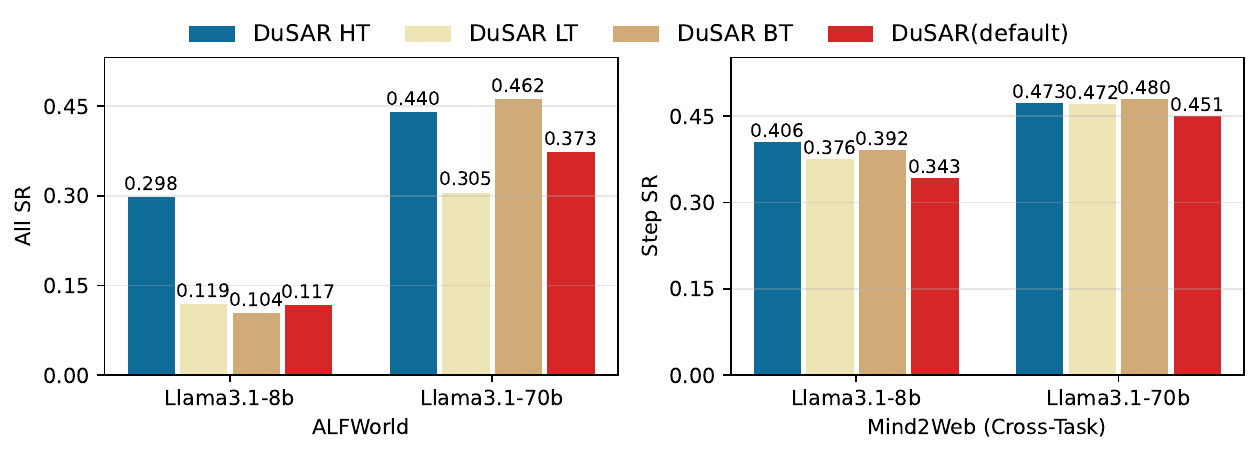}
\caption{Ablation results of expert demonstration application on holistic (DuSAR HT), local (DuSAR LT), and bidirectional integration (DuSAR BT) on ALFWorld (All SR) and Mind2Web Cross-Task (Step SR).}
\label{fig:tae_effect}
\end{figure}

As shown in Figure~\ref{fig:tae_effect}, the effectiveness of demonstrations depends critically on model scale and the granularity of integration. On ALFWorld, high-level guidance (\textbf{HT}) provides substantial gains for smaller models: Llama3.1-8B achieves 29.8\% SR (+18.1 percentage points over default DuSAR), as global structure compensates for limited planning capacity. In contrast, low-level traces (\textbf{LT}/\textbf{BT}) offer limited benefit at this scale (11.9\%/10.4\% SR), likely because fine-grained action sequences are difficult to contextualize without sufficient reasoning bandwidth. On ALFWorld with Llama3.1-70B, however, \textbf{BT} achieves 46.2\% SR (+8.9 percentage points), outperforming even HT (44.0\% SR). This demonstrates that large models can effectively integrate multi-granular demonstrations when reasoning capacity is abundant.

On Mind2Web, high-level guidance (\textbf{HT}) yields the highest Step SR for 8B (40.6\% vs. 37.6\% for LT and 34.3\% for default), providing a modest 6.3\% gain. All variants converge near 47--48\% Step SR at 70B, yet default DuSAR already achieves 45.1\% Step SR (Cross-Task) and 4.00\% Task SR (All) without any demonstrations. This demonstrates that while DuSAR's Step SR of 45.1\% measures per-step action correctness, its Task SR of 4.00\% represents the more stringent metric of complete task completion, achieving a 2.0$\times$ improvement over TRAD's 1.75\% Task SR. 

\paragraph{Reduced Dependence on Demonstrations}
These results establish that while external traces can enhance performance in specific regimes—particularly high-level guidance for small models or multi-granular fusion for large ones—DuSAR’s default zero-shot configuration remains highly competitive. It matches or approaches the best augmented variants across settings, especially under resource constraints. This confirms that a well-designed inductive bias—global decomposition coupled with local correction—is not merely a fallback in the absence of demonstrations, but a robust and efficient foundation for generalizable planning that scales effectively without external supervision.

\section{Conclusion}
\label{sec:conclusion}

We present DuSAR, a demonstration-free, training-free framework that enables robust long-horizon planning in frozen open-weight LLMs through co-adaptive dual-strategy reasoning, which combines holistic subgoal decomposition with local, observation-grounded execution. DuSAR significantly outperforms retrieval-augmented baselines across simulation and web environments, all without demonstrations or fine-tuning. Crucially, this approach empowers even small models to generalize across domains via zero-shot abstraction, revealing that effective planning stems from architectural inductive bias rather than model capacity alone.

\bibliography{dusar}

\section{Appendix}
\subsection{Prompt Templates}
\label{app:prompts}

Due to space constraints, we provide key prompt templates here. Full templates are available in the code repository. We illustrate template variations across different task scenarios to demonstrate how DuSAR adapts its reasoning structure to domain-specific requirements.

\subsubsection{Core Prompt Templates}

\textbf{Holistic Reflect Prompt Template (Base):}
\begin{lstlisting}[basicstyle=\small\ttfamily,breaklines=true,numbers=none,frame=none]
You are a strategic planner. Given a task instruction, past exploration 
history, and current holistic strategy, generate a refined high-level plan.

Task: {task_instruction}
Previous Strategy: {H_{t-1}}
Exploration History: {E_{<t}}
Strategy Fitness Score: {s_{t-1}}

Generate a holistic strategy that:
1. Decomposes the task into ordered sub-goals
2. Accounts for progress indicated by the fitness score
3. Revises the plan if stagnation (score=0) or advances if milestone 
   achieved (50 <= score <= 99)
\end{lstlisting}

\textbf{Local Reflect Prompt Template (Base):}
\begin{lstlisting}[basicstyle=\small\ttfamily,breaklines=true,numbers=none,frame=none]
You are a local executor. Given current observation, holistic strategy, 
and exploration history, generate immediate action guidance.

Observation: {o_t}
Holistic Strategy: {H_t}
Exploration History: {E_{<t}}

Generate a local strategy that:
1. Identifies feasible next actions
2. Evaluates alignment with holistic goals
3. Provides context-aware execution guidance
\end{lstlisting}

\textbf{Score Analysis Prompt Template (Base):}
\begin{lstlisting}[basicstyle=\small\ttfamily,breaklines=true,numbers=none,frame=none]
Evaluate the Strategy Fitness Score based on:
- Current observation: {o_t}
- Last action: {a_{t-1}}
- Reward signal: {r_t}
- Recent history: {E_{<t}}

Assign numerical score (0-100):
- 0: No progress (stagnation, repeated actions, no state change)
- 1-49: Approaching sub-goal (relevant elements found, ongoing exploration)
- 50-99: Significant advancement or sub-goal completed (milestone achieved)
- 100: Task completed (overall goal reached)
\end{lstlisting}

\subsubsection{ALFWorld Scenario-Specific Templates}

DuSAR adapts its prompts based on ALFWorld task types. Below are examples for each scenario:

\textbf{1. Put Task (pick\_and\_place):} "put apple in fridge"
\begin{itemize}
    \item \textit{Holistic Strategy Focus:} "Decompose: (1) Locate apple, (2) Pick up apple, (3) Navigate to fridge, (4) Place apple in fridge"
    \item \textit{Local Strategy Emphasis:} Object location identification, container state checking (open/closed), navigation planning
    \item \textit{Score Interpretation:} Milestone at 50 when object found, 75 when picked up, 100 when placed
\end{itemize}

\textbf{2. Examine Task (look\_at\_obj):} "look at bowl under desklamp"
\begin{itemize}
    \item \textit{Holistic Strategy Focus:} "Decompose: (1) Locate bowl, (2) Locate desklamp, (3) Position bowl under desklamp, (4) Use desklamp"
    \item \textit{Local Strategy Emphasis:} Spatial relationship understanding, object positioning, tool usage
    \item \textit{Score Interpretation:} Milestone at 50 when both objects found, 75 when positioned correctly, 100 when desklamp used
\end{itemize}

\textbf{3. Clean Task (pick\_clean\_then\_place):} "clean egg then put in microwave"
\begin{itemize}
    \item \textit{Holistic Strategy Focus:} "Decompose: (1) Locate egg, (2) Pick up egg, (3) Navigate to sinkbasin, (4) Clean egg, (5) Navigate to microwave, (6) Place egg"
    \item \textit{Local Strategy Emphasis:} Multi-step sequence tracking, operation-specific locations (sinkbasin for cleaning), state transitions
    \item \textit{Score Interpretation:} Milestone at 50 when egg picked, 75 when cleaned, 90 when placed, 100 when task complete
\end{itemize}

\textbf{4. Heat Task (pick\_heat\_then\_place):} "heat mug then put in garbagecan"
\begin{itemize}
    \item \textit{Holistic Strategy Focus:} "Decompose: (1) Locate mug, (2) Pick up mug, (3) Navigate to microwave, (4) Heat mug, (5) Navigate to garbagecan, (6) Place mug"
    \item \textit{Local Strategy Emphasis:} Temperature operation sequencing, microwave state management, post-operation navigation
    \item \textit{Score Interpretation:} Milestone at 50 when mug picked, 75 when heated, 90 when placed, 100 when task complete
\end{itemize}

\textbf{5. Cool Task (pick\_cool\_then\_place):} "cool potato then put in cabinet"
\begin{itemize}
    \item \textit{Holistic Strategy Focus:} "Decompose: (1) Locate potato, (2) Pick up potato, (3) Navigate to fridge, (4) Cool potato, (5) Navigate to cabinet, (6) Place potato"
    \item \textit{Local Strategy Emphasis:} Cooling operation sequencing, fridge state management, temperature-sensitive placement
    \item \textit{Score Interpretation:} Milestone at 50 when potato picked, 75 when cooled, 90 when placed, 100 when task complete
\end{itemize}

\textbf{6. PutTwo Task (pick\_two\_obj):} "put two soapbars in garbagecan"
\begin{itemize}
    \item \textit{Holistic Strategy Focus:} "Decompose: (1) Locate first soapbar, (2) Pick up first soapbar, (3) Locate second soapbar, (4) Pick up second soapbar, (5) Navigate to garbagecan, (6) Place both soapbars"
    \item \textit{Local Strategy Emphasis:} Multi-object tracking, progress monitoring for each object, preventing redundant exploration
    \item \textit{Score Interpretation:} Milestone at 25 when first object found, 50 when first picked, 75 when second found, 90 when second picked, 100 when both placed
\end{itemize}

\subsubsection{Mind2Web Scenario-Specific Templates}

DuSAR adapts prompts for web navigation tasks across three generalization settings:

\textbf{1. Cross-Task Setting:} Same website, different task types
\begin{itemize}
    \item \textit{Example Task:} "Book a hotel room in Paris" (booking website)
    \item \textit{Holistic Strategy Focus:} "Decompose: (1) Locate booking form, (2) Fill destination field, (3) Fill date fields, (4) Fill guest information, (5) Submit booking"
    \item \textit{Local Strategy Emphasis:} Form field identification, input validation, sequential form completion
    \item \textit{Score Interpretation:} Milestone at 30 when form located, 50 when destination filled, 75 when dates filled, 90 when all fields complete, 100 when submitted
    \item \textit{Template Adaptation:} Emphasizes form structure understanding and field dependencies
\end{itemize}

\textbf{2. Cross-Website Setting:} Different websites, similar task types
\begin{itemize}
    \item \textit{Example Task:} "Search for romantic reggae music from BCD Studio for TikTok in Andorra" (music discovery website)
    \item \textit{Holistic Strategy Focus:} "Decompose: (1) Locate search interface, (2) Enter search terms (genre, artist, location), (3) Filter results, (4) Select appropriate content"
    \item \textit{Local Strategy Emphasis:} Website-specific UI patterns, search interface adaptation, result filtering
    \item \textit{Score Interpretation:} Milestone at 25 when search interface found, 50 when search executed, 75 when results filtered, 100 when content selected
    \item \textit{Template Adaptation:} Emphasizes cross-website UI pattern recognition and adaptive interaction
\end{itemize}

\textbf{3. Cross-Domain Setting:} Different domains, diverse task types
\begin{itemize}
    \item \textit{Example Task:} "Find pill with imprint M366" (medical/pharmacy website)
    \item \textit{Holistic Strategy Focus:} "Decompose: (1) Locate search/identification tool, (2) Enter pill identifier (M366), (3) Review identification results, (4) Verify pill information"
    \item \textit{Local Strategy Emphasis:} Domain-specific terminology, specialized interface elements, information verification
    \item \textit{Score Interpretation:} Milestone at 20 when search tool found, 40 when identifier entered, 60 when results displayed, 80 when information verified, 100 when task complete
    \item \textit{Template Adaptation:} Emphasizes domain knowledge adaptation and specialized interface handling
\end{itemize}

\subsubsection{Template Adaptation Mechanism}

The core templates remain consistent across scenarios, but DuSAR dynamically adjusts emphasis based on:
\begin{itemize}
    \item \textbf{Task Complexity:} PutTwo (6 steps) receives more granular milestone tracking than Put (4 steps)
    \item \textbf{Domain Characteristics:} ALFWorld emphasizes spatial navigation and object manipulation; Mind2Web emphasizes UI element interaction and form completion
    \item \textbf{Generalization Setting:} Cross-Domain requires more adaptive terminology handling than Cross-Task
    \item \textbf{Progress Indicators:} Score thresholds adapt to task length—longer tasks use more intermediate milestones (25, 50, 75, 90) while shorter tasks use fewer (50, 75, 100)
\end{itemize}

This adaptive template mechanism enables DuSAR to maintain consistent reasoning structure while accommodating diverse task requirements across benchmarks and generalization settings.

\subsection{Explore Trace Example}
\label{app:trace}

Table~\ref{tab:trace_example} shows a complete Explore Trace from an ALFWorld task "put some saltshaker on cabinet", demonstrating how DuSAR's co-adaptive loop works with numerical Strategy Fitness Scores. Due to space constraints, we provide a simplified version here; full traces are available in the code repository.

\begin{table}[htbp]
\footnotesize
\centering
\caption{Explore Trace Example from ALFWorld Task (simplified)}
\label{tab:trace_example}
\setlength{\tabcolsep}{2pt}
\begin{tabular}{c p{2.6cm} p{1.7cm} c p{2.3cm}}
\toprule
Step & Observation & Action & Score & Reasoning \\
\midrule
1 & Loc 22, ctop 2: apple, bowl... & go ctop 2 & 25 & Exploring \\
\midrule
2 & Loc 12, ctop 1: mug, soapbottle & go ctop 1 & 25 & Continuing \\
\midrule
3 & Loc 5, ctop 3: ... & go ctop 3 & 25 & Exploring (1--49) \\
\midrule
4 & Loc 13, cab 1: glassbottle & go cab 1 & 25 & Expanding \\
\midrule
5 & Loc 11, cab 2: closed & go cab 2 & 25 & Found closed \\
\midrule
6 & Opened cab 2: cup & open cab 2 & 50 & Opened \\
\bottomrule
\end{tabular}
\end{table}

\textbf{Score Interpretation:}
\begin{itemize}
    \item \textbf{Score 25} (Steps 1--5): Ongoing exploration, no saltshaker found yet. Score falls in range $1 \leq s_t \leq 49$, so Holistic Strategy is maintained.
    \item \textbf{Score 50} (Step 6): Significant progress—cabinet opened. Score in range $50 \leq s_t \leq 99$, triggering Holistic Strategy refinement to focus on remaining cabinets.
\end{itemize}

This trace illustrates how the numerical score ($0$--$100$) provides fine-grained feedback, enabling the co-adaptive loop to distinguish between ongoing progress (maintain strategy) and significant milestones (update strategy).

\begin{table*}[htbp]
\footnotesize
\centering
\caption{Comparison of DuSAR with Baseline Methods}
\label{tab:method_comparison}
\setlength{\tabcolsep}{3pt}
\begin{tabular}{l p{2cm} p{3cm} p{2.3cm} p{2.2cm}}
\toprule
Method & Demons. & Planning Approach & Adaptation & Token Efficiency \\
\midrule
\textbf{ReAct}~\cite{yao2023react} & \ding{51} (fixed) & Sequential 

reasoning-acting loop & Reactive      

self-correction & Medium 

(1.4k--1.9k/step) \\
\midrule
\textbf{Synapse}~\cite{zheng2024synapse} & \ding{51} (trajectory) & Single-stage 

trajectory retrieval & Task-based 

exemplar 

selection & Moderate 

(1.5k--2.1k/step) \\
\midrule
\textbf{TRAD}~\cite{zhou2024trad} & \ding{51} (step-wise) & Two-stage retrieval 

(task + thought) & Thought-based 

dynamic 

refinement & Low 

(3.2k--3.6k/step) \\
\midrule
\textbf{DuSAR} (Ours) & \ding{55} & Co-adaptive 

dual-strategy & Dynamic 

co-adaptation & High 

(333--440/step) \\
\bottomrule
\end{tabular}
\end{table*}

\paragraph{Mind2Web Trace Example.} 
For web navigation tasks, Explore Traces capture HTML observations, action selections, and scores similarly. A typical trace entry from Mind2Web includes: \texttt{observation} (simplified DOM with top-5 ranked elements), \texttt{action} (e.g., \texttt{CLICK [142]} for clicking pickup location field), \texttt{score} (e.g., 75 when relevant element is successfully identified), and \texttt{local\_log} (reasoning for action selection). The Holistic Strategy updates when scores indicate significant progress (e.g., form field filled, $s_t \geq 50$) or stagnation (e.g., repeated invalid clicks, $s_t = 0$).

In contrast, ReAct's traces lack structured strategy evolution: each step generates independent reasoning without global coordination, leading to inefficient exploration patterns. For instance, ReAct may repeatedly click similar elements without recognizing task completion milestones, resulting in lower task success rates (0.13\% vs. DuSAR's 4.00\% on Mind2Web with Llama3.1-70B).

Full trace schemas and additional examples are available in the code repository.

\subsection{Comparison with Baseline Methods}
\label{app:comparison}

Table~\ref{tab:method_comparison} provides a systematic comparison of DuSAR against representative baseline methods. Below, we illustrate their differences through a concrete example.

\paragraph{Case Study: ALFWorld PutTwo Task "Put two soapbars in garbagecan"}

We illustrate how each method handles this complex 6-step task requiring sequential object manipulation:

\textbf{ReAct (Llama3.1-70B, 0\% success rate):}
\begin{itemize}
    \item \textit{Step 1:} "Thought: I need to find soapbars. Let me explore..." → Action: \texttt{go to cabinet 1}
    \item \textit{Step 2:} "Thought: No soapbars here. Let me check another location..." → Action: \texttt{go to countertop 2}
    \item \textit{Step 3:} "Thought: Still no soapbars..." → Action: \texttt{go to cabinet 3}
    \item \textit{Step 4:} "Thought: Found one soapbar. I should pick it up..." → Action: \texttt{pick soapbar 1}
    \item \textit{Step 5:} "Thought: Now I need another soapbar..." → Action: \texttt{go to cabinet 4} (revisits cabinet 1, losing track)
    \item \textit{Step 6:} "Thought: Let me check cabinet 1 again..." → Action: \texttt{go to cabinet 1} (redundant exploration)
    \item \textit{Failure Analysis:} No global plan; each step's reasoning is independent. After finding first soapbar, ReAct cannot systematically track that it needs a \textit{second} soapbar, leading to redundant exploration of previously visited locations. Achieves 0\% success rate across all model scales.
\end{itemize}

\textbf{Synapse (Llama3.1-70B, 0\% success rate):}
\begin{itemize}
    \item \textit{Step 1:} Single-stage retrieval: Retrieve exemplar "Task: put apple in fridge. Steps: find apple → pick apple → go to fridge → put apple" (3-step trajectory)
    \item \textit{Step 2:} Generate action following exemplar: \texttt{go to cabinet 1} (matching "find apple" pattern)
    \item \textit{Step 3:} Continue following exemplar: \texttt{pick soapbar 1} (matching "pick apple" pattern)
    \item \textit{Step 4:} Exemplar ends here—no guidance for second object. Agent attempts: \texttt{go to garbagecan} (premature placement)
    \item \textit{Failure Analysis:} Retrieved exemplar has only 3 steps, but PutTwo requires 6 steps. Synapse cannot extend beyond the exemplar's length, leading to premature task completion attempt. Single-stage retrieval lacks dynamic refinement to adapt to longer sequences.
\end{itemize}

\textbf{TRAD (Llama3.1-70B, 0\% success rate):}
\begin{itemize}
    \item \textit{Step 1:} First-stage retrieval: Retrieve exemplars based on task similarity "put two objects"
    \item \textit{Step 2:} Generate thought: "I need to find two soapbars and place them in garbagecan"
    \item \textit{Step 3:} Second-stage retrieval: Re-retrieve based on thought similarity—finds exemplars with "find object" reasoning
    \item \textit{Step 4:} Generate action: \texttt{go to cabinet 1} (from thought-aligned exemplar)
    \item \textit{Step 5:} Generate thought: "Found one soapbar, need to pick it up"
    \item \textit{Step 6:} Second-stage retrieval fails—thought "need second soapbar" has no matching exemplar traces in bank
    \item \textit{Failure Analysis:} Two-stage retrieval helps initially, but when thought-based retrieval cannot find relevant traces for multi-object scenarios, TRAD falls back to mismatched exemplars. Higher token cost (3.2k--3.6k/step) also limits context window on smaller models.
\end{itemize}

\textbf{DuSAR (Llama3.1-70B, 52.9\% success rate):}
\begin{itemize}
    \item \textit{Holistic Strategy (Step 1):} "Decompose: (1) Locate first soapbar, (2) Pick up first soapbar, (3) Locate second soapbar, (4) Pick up second soapbar, (5) Navigate to garbagecan, (6) Place both soapbars"
    \item \textit{Local Strategy (Step 1):} "Observe: cabinet 1 contains mug. Action: continue searching" → Score: $s_1 = 25$ (ongoing, $1 \leq s_1 \leq 49$) → Maintain $H_1$
    \item \textit{Local Strategy (Step 4):} "Found soapbar 1. Action: pick soapbar 1" → Score: $s_4 = 75$ (milestone, $50 \leq s_4 \leq 99$) → Update $H_4$: "Progress: soapbar 1 located and picked. Next: find second soapbar, avoid revisiting cabinet 1"
    \item \textit{Local Strategy (Step 7):} "Found soapbar 2. Action: pick soapbar 2" → Score: $s_7 = 90$ (milestone) → Update $H_7$: "Progress: both soapbars acquired. Next: navigate to garbagecan"
    \item \textit{Success Factor:} Co-adaptive refinement enables dynamic plan adjustment. When first soapbar is found ($s_4 = 75$), Holistic Strategy updates to explicitly track the second object requirement, preventing redundant exploration. Score milestones (25, 50, 75, 90, 100) provide fine-grained progress tracking for the 6-step sequence.
\end{itemize}

This case study demonstrates how DuSAR's dual-strategy framework enables efficient multi-step planning through hierarchical decomposition and adaptive refinement. ReAct lacks global coordination, leading to inefficient exploration patterns and 0\% success rate on PutTwo, even with fixed hardcoded examples. Retrieval-based methods (Synapse, TRAD) fail when exemplars are mismatched or incomplete, also achieving 0\% on PutTwo across all model scales. DuSAR's internal reasoning and co-adaptive loop enable 52.9\% success rate without any demonstrations.

\paragraph{Case Study: Mind2Web Cross-Task Setting "View top-rated new restaurants in Toronto and book a table for two"}

\textbf{ReAct (Llama3.1-70B, 0\% success rate):}
\begin{itemize}
    \item \textit{Step 1:} Uses fixed exemplars: Selects first N exemplars from memory (2 fixed exemplars, no dynamic similarity matching). The exemplars are pre-selected trajectories from the training set, not task-specific hardcoded examples. Thought: "I need to find restaurants..." → Action: \texttt{CLICK [Toronto]} (correct initial step)
    \item \textit{Step 2:} Thought: "Now I need to select party size..." → Action: \texttt{SELECT [70027] [2 Guests]} (correct, but following fixed exemplar pattern)
    \item \textit{Step 3:} Thought: "I should search for Malaysian restaurants..." → Action: \texttt{TYPE [76848] [Malaysian restaurants in Toronto]} (incorrect—should click Reserve button, not search again)
    \item \textit{Step 4:} Thought: "I think I'm done..." → Action: \texttt{SELECT [76948] [2]} (redundant action, missing final confirmation)
    \item \textit{Failure Analysis:} Fixed exemplars selected from memory do not match restaurant booking task structure. ReAct follows exemplar patterns (sequential navigation) but cannot adapt to task-specific requirements (final confirmation step). No structured plan for multi-step booking workflows. Each step's reasoning is independent, leading to missed critical steps (e.g., final reservation confirmation). Token cost: 1.4k--1.9k per step.
\end{itemize}

\textbf{Synapse (Llama3.1-70B, 0.40\% success rate):}
\begin{itemize}
    \item \textit{Step 1:} Single-stage retrieval: Retrieve exemplar "Task: Find gift registry. Steps: CLICK [Gift Registry] → CLICK [Find Registry] → TYPE [First Name] → TYPE [Last Name]" (4-step trajectory)
    \item \textit{Step 2:} Generate action following exemplar: \texttt{CLICK [booking link]} (matching "CLICK [Gift Registry]" pattern)
    \item \textit{Step 3:} Continue following exemplar: \texttt{TYPE [destination] [Paris]} (matching "TYPE [First Name]" pattern)
    \item \textit{Step 4:} Exemplar pattern mismatch—booking forms require date fields, but exemplar only shows name fields. Agent attempts: \texttt{CLICK [submit]} (premature submission, missing required date/guest fields)
    \item \textit{Failure Analysis:} Single-stage retrieval based on task similarity retrieves mismatched exemplars (gift registry vs. hotel booking). Synapse cannot adapt when exemplar structure differs from current task requirements, leading to incomplete form submissions. Token cost: 1.5k--2.1k per step.
\end{itemize}

\textbf{TRAD (Llama3.1-70B, 1.60\% success rate):}
\begin{itemize}
    \item \textit{Step 1:} First-stage retrieval: Retrieve exemplars based on task similarity "book hotel" → finds booking-related trajectories
    \item \textit{Step 2:} Generate thought: "I need to fill the booking form with destination, dates, and guest information"
    \item \textit{Step 3:} Second-stage retrieval: Re-retrieve based on thought similarity—finds exemplars with "fill form" reasoning
    \item \textit{Step 4:} Generate action: \texttt{TYPE [destination] [Paris]} (from thought-aligned exemplar)
    \item \textit{Step 5:} Generate thought: "Destination filled, now need dates"
    \item \textit{Step 6:} Second-stage retrieval finds exemplars with date-filling reasoning, enabling: \texttt{CLICK [date picker]}
    \item \textit{Step 7:} Thought "need guest count" has no matching exemplar—falls back to mismatched exemplar, skips guest field
    \item \textit{Failure Analysis:} Two-stage retrieval improves exemplar matching initially, but when thought-based retrieval cannot find relevant traces for specific form fields (e.g., guest count), TRAD falls back to incomplete exemplars. Higher token cost (3.2k--3.6k/step) limits context window on smaller models.
\end{itemize}

\textbf{DuSAR (Llama3.1-70B, 5.28\% success rate):}
\begin{itemize}
    \item \textit{Holistic Strategy (Step 1):} "Decompose: (1) Locate booking form, (2) Fill destination field, (3) Fill date fields, (4) Fill guest information, (5) Submit booking"
    \item \textit{Local Strategy (Step 2):} "Observe: destination field [19054] visible. Action: TYPE [19054] [Paris]" → Score: $s_2 = 30$ (ongoing, $1 \leq s_2 \leq 49$) → Maintain $H_2$
    \item \textit{Local Strategy (Step 3):} "Destination filled. Action: CLICK [date picker]" → Score: $s_3 = 50$ (milestone) → Update $H_3$: "Progress: destination complete. Next: fill dates and guests"
    \item \textit{Local Strategy (Step 5):} "All required fields filled. Action: CLICK [submit]" → Score: $s_5 = 80$ (milestone) → Update $H_5$: "Progress: form completed. Verify confirmation"
    \item \textit{Success Factor:} Structured decomposition ensures all required fields are systematically addressed. Score milestones (30, 50, 80, 100) track form completion progress, preventing skipped fields. Token cost: 335--564 per step (To\_Pro), 3--4× lower than retrieval methods.
\end{itemize}

\paragraph{Case Study: Mind2Web Cross-Website Setting "Find a used cheapest 8 cylinder BMW made between 2005-2015"}

\textbf{ReAct (Llama3.1-70B, 0\% success rate):}
\begin{itemize}
    \item \textit{Step 1:} Uses fixed exemplars: Selects first N exemplars from memory (2 fixed exemplars, no dynamic similarity matching). The exemplars are pre-selected trajectories from the training set, not task-specific hardcoded examples. Thought: "I need to select used cars..." → Action: \texttt{SELECT [328] [Used]} (correct initial step)
    \item \textit{Step 2:} Thought: "Now I need to select BMW..." → Action: \texttt{SELECT [8508] [BMW]} (correct, following exemplar pattern)
    \item \textit{Step 3:} Thought: "I should set price range..." → Action: \texttt{SELECT [21320] [\$25,000]} and \texttt{SELECT [55685] [\$50,000]} (correct)
    \item \textit{Step 4:} Thought: "I need to select year range..." → Action: \texttt{SELECT [91167] [2005]} and \texttt{SELECT [126457] [2015]} (correct)
    \item \textit{Step 5:} Thought: "I should select cylinders..." → Action: \texttt{CLICK [175212]} (Cylinders button, correct) then \texttt{CLICK [175251]} (8 cylinders checkbox, correct)
    \item \textit{Step 6:} Thought: "I need to set mileage..." → Action: \texttt{CLICK [208583]} (wrong—should select mileage dropdown, not cylinders again)
    \item \textit{Step 7:} Thought: "I should sort by price..." → Action: \texttt{SELECT [262383] [Lowest price]} (correct, but task incomplete—missing mileage filter)
    \item \textit{Failure Analysis:} Fixed exemplars selected from memory do not match car search task structure. ReAct follows exemplar patterns (sequential selection) but cannot adapt to website-specific UI patterns (different filter layouts across car websites). Repeated errors on mileage filter indicate inability to learn from feedback. No structured approach to understanding different website layouts. Token cost: 1.4k--1.9k per step.
\end{itemize}

\textbf{Synapse (Llama3.1-70B, 0.00\% success rate):}
\begin{itemize}
    \item \textit{Step 1:} Single-stage retrieval: Retrieve exemplar "Task: Find Hot Dogs in Oakland. Steps: TYPE [Find] [Hot Dogs] → CLICK [Hot Dogs]" (2-step trajectory from different website)
    \item \textit{Step 2:} Generate action following exemplar: \texttt{TYPE [51560] [romantic reggae]} (matching "TYPE [Find]" pattern, but wrong field format)
    \item \textit{Step 3:} Website-specific UI differs—exemplar's search interface structure doesn't match current website. Agent attempts: \texttt{CLICK [wrong element]} (UI mismatch)
    \item \textit{Failure Analysis:} Single-stage retrieval retrieves exemplars from different websites with different UI patterns. Synapse cannot adapt when exemplar's UI structure (field IDs, layout) differs from current website, leading to incorrect element selection. Token cost: 1.5k--2.1k per step.
\end{itemize}

\textbf{TRAD (Llama3.1-70B, 1.10\% success rate):}
\begin{itemize}
    \item \textit{Step 1:} First-stage retrieval: Retrieve exemplars based on task similarity "search music" → finds music search trajectories
    \item \textit{Step 2:} Generate thought: "I need to search for reggae music from BCD Studio in Andorra"
    \item \textit{Step 3:} Second-stage retrieval: Re-retrieve based on thought similarity—finds exemplars with "search music" reasoning, but from different websites
    \item \textit{Step 4:} Generate action: \texttt{TYPE [51560] [romantic reggae BCD Studio Andorra]} (from thought-aligned exemplar, but UI format may differ)
    \item \textit{Step 5:} Website-specific UI adaptation fails—thought-based retrieval cannot account for cross-website UI differences (different field IDs, search formats)
    \item \textit{Failure Analysis:} Two-stage retrieval improves semantic matching, but cannot handle cross-website UI pattern differences. When exemplars come from websites with different UI structures, TRAD's thought-based matching still produces mismatched actions. Higher token cost (3.2k--3.6k/step) limits context for UI pattern learning.
\end{itemize}

\textbf{DuSAR (Llama3.1-70B, 3.11\% success rate):}
\begin{itemize}
    \item \textit{Holistic Strategy (Step 1):} "Decompose: (1) Locate car search interface, (2) Select filters (used cars, BMW, 8 cylinders, year range 2005-2015), (3) Sort by price (lowest), (4) Select appropriate vehicle"
    \item \textit{Local Strategy (Step 1):} "Observe: filter interface found. Action: SELECT [328] [Used]" → Score: $s_1 = 25$ (ongoing) → Maintain $H_1$
    \item \textit{Local Strategy (Step 2):} "Used filter selected. Action: SELECT [8508] [BMW]" → Score: $s_2 = 30$ (ongoing) → Maintain $H_2$
    \item \textit{Local Strategy (Step 5):} "All filters applied. Action: SELECT [262383] [Lowest price]" → Score: $s_5 = 75$ (milestone) → Update $H_5$: "Progress: filters complete. Next: select vehicle"
    \item \textit{Success Factor:} Cross-website UI pattern recognition enables adaptation to different search interfaces. Holistic Strategy provides flexible decomposition that adapts to website-specific layouts without relying on exemplar UI structures. Token cost: 335--564 per step (To\_Pro).
\end{itemize}

\paragraph{Case Study: Mind2Web Cross-Domain Setting "Show me the page with information about Adderall side effects"}

\textbf{ReAct (Llama3.1-70B, 0.38\% success rate):}
\begin{itemize}
    \item \textit{Step 1:} Uses fixed exemplars: Selects first N exemplars from memory (2 fixed exemplars, no dynamic similarity matching). The exemplars are pre-selected trajectories from the training set, not task-specific hardcoded examples. Thought: "I need to find Adderall side effects..." → Action: \texttt{CLICK [153]} (Side Effects link, correct initial step)
    \item \textit{Step 2:} Thought: "I should click on Adderall to get more information..." → Action: \texttt{CLICK [5394]} (Adderall link, but wrong context—should stay on side effects page)
    \item \textit{Step 3:} Thought: "I'm confused about the interface..." → Action: Outputs observation HTML instead of action (format error)
    \item \textit{Failure Analysis:} Fixed exemplars selected from memory do not match medical/pharmacy domain structure. ReAct follows exemplar patterns (sequential clicking) but cannot adapt to domain-specific requirements (staying on specialized pages vs. general navigation). Lacks understanding of specialized medical interfaces and terminology. No structured approach to domain knowledge adaptation. Token cost: 1.4k--1.9k per step.
\end{itemize}

\textbf{Synapse (Llama3.1-70B, 0.70\% success rate):}
\begin{itemize}
    \item \textit{Step 1:} Single-stage retrieval: Retrieve exemplar "Task: Find dog food. Steps: TYPE [search] [dog food] → CLICK [result]" (general search trajectory, not domain-specific)
    \item \textit{Step 2:} Generate action following exemplar: \texttt{TYPE [4829] [M366]} (matching "TYPE [search]" pattern, but wrong field—imprint requires specialized tool, not general search)
    \item \textit{Step 3:} Domain mismatch—exemplar from e-commerce (general search) doesn't match medical/pharmacy domain (specialized imprint identification tool). Agent attempts: \texttt{CLICK [wrong element]} (domain-specific UI not recognized)
    \item \textit{Failure Analysis:} Single-stage retrieval based on task similarity retrieves exemplars from different domains (e-commerce vs. medical). Synapse cannot adapt when exemplar domain differs from current task domain, leading to incorrect tool selection (general search vs. specialized imprint tool). Token cost: 1.5k--2.1k per step.
\end{itemize}

\textbf{TRAD (Llama3.1-70B, 1.50\% success rate):}
\begin{itemize}
    \item \textit{Step 1:} First-stage retrieval: Retrieve exemplars based on task similarity "find item" → finds general search trajectories from e-commerce domains
    \item \textit{Step 2:} Generate thought: "I need to find a pill with imprint M366"
    \item \textit{Step 3:} Second-stage retrieval: Re-retrieve based on thought similarity—finds exemplars with "search" reasoning, but from wrong domain (e-commerce search vs. medical imprint identification)
    \item \textit{Step 4:} Generate action: \texttt{TYPE [4829] [M366]} (from thought-aligned exemplar, but wrong domain—should use specialized imprint tool, not general search)
    \item \textit{Step 5:} Domain-specific adaptation fails—thought-based retrieval cannot distinguish between general search (e-commerce) and specialized identification tools (medical/pharmacy)
    \item \textit{Failure Analysis:} Two-stage retrieval improves semantic matching within similar domains, but fails under cross-domain distribution shift. When exemplars come from different domains (e-commerce vs. medical), TRAD's thought-based matching produces domain-mismatched actions. Higher token cost (3.2k--3.6k/step) limits context for domain-specific pattern learning.
\end{itemize}

\textbf{DuSAR (Llama3.1-70B, 3.61\% success rate):}
\begin{itemize}
    \item \textit{Holistic Strategy (Step 1):} "Decompose: (1) Locate medication search/navigation interface, (2) Navigate to Adderall information page, (3) Locate side effects section, (4) Verify side effects information is displayed"
    \item \textit{Local Strategy (Step 1):} "Observe: Side Effects link [153] found. Action: CLICK [153]" → Score: $s_1 = 30$ (ongoing) → Maintain $H_1$
    \item \textit{Local Strategy (Step 2):} "Side effects page loaded. Action: navigate to Adderall section" → Score: $s_2 = 50$ (milestone) → Update $H_2$: "Progress: navigation complete. Next: locate side effects information"
    \item \textit{Local Strategy (Step 4):} "Side effects information displayed. Action: verify content" → Score: $s_4 = 100$ (milestone) → Update $H_4$: "Progress: task complete. Side effects information verified"
    \item \textit{Success Factor:} Domain knowledge adaptation enables handling of specialized terminology and interfaces. Holistic Strategy decomposes medical/pharmacy tasks into domain-appropriate steps, while Local Strategy adapts to specialized UI elements without relying on domain-matched exemplars. Token cost: 335--564 per step (To\_Pro), enabling efficient cross-domain generalization.
\end{itemize}

These case studies demonstrate how DuSAR's adaptive template mechanism enables consistent reasoning structure while accommodating diverse task requirements. Cross-Task benefits from form structure understanding, Cross-Website from UI pattern recognition, and Cross-Domain from specialized terminology handling—all within the same dual-strategy framework.

\textbf{Key Architectural Differences:}

\begin{itemize}
    \item \textbf{Demonstration Requirements:} Unlike Synapse (single-stage trajectory-level retrieval) and TRAD (two-stage step-level retrieval with thought-based refinement), DuSAR operates demonstration-free, enabling zero-shot generalization. ReAct~\cite{yao2023react} uses fixed hardcoded examples (2 per task type) rather than dynamic retrieval, but still relies on external demonstrations. ReAct's simple reasoning-acting loop struggles on open-source LLMs even with fixed examples: 1.49\% SR on ALFWorld (Llama3.1-70B) vs. DuSAR's 37.3\%, and 0.13\% SR on Mind2Web (Llama3.1-70B) vs. DuSAR's 4.00\%.
    
    \item \textbf{Planning Architecture:} DuSAR's dual-strategy (holistic-local) framework enables hierarchical plan decomposition and dynamic refinement through co-adaptive feedback. Synapse uses single-stage retrieval to match full trajectories based on task similarity, generating actions directly from exemplars. TRAD employs a two-stage approach: first retrieving exemplars by task similarity, then re-retrieving by thought similarity after generating intermediate reasoning, enabling more precise exemplar matching. TRAD also supports multi-step context windows and historical step integration for enhanced contextual awareness. However, both retrieval-based methods match fixed exemplars from external banks, lacking internal plan generation. ReAct uses sequential step-by-step reasoning without global coordination or hierarchical structure.
    
    \item \textbf{Adaptation Mechanism:} DuSAR employs a co-adaptive loop where both strategies iteratively refine each other based on Strategy Fitness Scores ($s_t \in [0, 100]$), enabling proactive plan advancement ($50 \leq s_t \leq 99$) and error recovery ($s_t = 0$). Synapse adapts only through initial task-based exemplar selection, with no dynamic refinement. TRAD adapts through two-stage retrieval (task similarity → thought similarity), providing more dynamic exemplar matching but still constrained by external demonstration banks. ReAct adapts reactively when actions fail, without structured strategy evolution.
    
    \item \textbf{Token Efficiency:} DuSAR achieves 3--9× token reduction (335--564 vs. 1.5k--3.7k per step for retrieval methods, To\_Pro) by eliminating demonstration overhead. Synapse incurs moderate token costs (1.5k--2.1k per step) due to single-stage trajectory retrieval. TRAD incurs higher token costs (3.2k--3.6k per step) due to two-stage retrieval (task similarity + thought similarity) and optional multi-step context expansion. ReAct uses moderate tokens (1.4k--2.2k per step) due to fixed hardcoded examples, but with limited performance on open-source models. DuSAR's efficiency advantage is particularly pronounced in long-horizon tasks where retrieval methods accumulate demonstration overhead across steps.
    
    \item \textbf{Generalization Capability:} DuSAR's internal reasoning enables strong cross-domain generalization, whereas retrieval-based methods degrade under distribution shift when exemplars are mismatched. ReAct generalizes well on proprietary models (GPT-4) but struggles on open-source backbones even with fixed examples, highlighting the importance of structured internal reasoning for resource-efficient deployment.
\end{itemize}

\setlength{\leftmargini}{20pt}
\makeatletter\def\@listi{\leftmargin\leftmargini \topsep .5em \parsep .5em \itemsep .5em}
\def\@listii{\leftmargin\leftmarginii \labelwidth\leftmarginii \advance\labelwidth-\labelsep \topsep .4em \parsep .4em \itemsep .4em}
\def\@listiii{\leftmargin\leftmarginiii \labelwidth\leftmarginiii \advance\labelwidth-\labelsep \topsep .4em \parsep .4em \itemsep .4em}\makeatother

\setcounter{secnumdepth}{0}
\renewcommand\thesubsection{\arabic{subsection}}
\renewcommand\labelenumi{\thesubsection.\arabic{enumi}}

\newcounter{checksubsection}
\newcounter{checkitem}[checksubsection]

\newcommand{\checksubsection}[1]{%
  \refstepcounter{checksubsection}%
  \paragraph{\arabic{checksubsection}. #1}%
  \setcounter{checkitem}{0}%
}

\newcommand{\checkitem}{%
  \refstepcounter{checkitem}%
  \item[\arabic{checksubsection}.\arabic{checkitem}.]%
}
\newcommand{\question}[2]{\normalcolor\checkitem #1 #2 \color{blue}}
\newcommand{\ifyespoints}[1]{\makebox[0pt][l]{\hspace{-15pt}\normalcolor #1}}

\bigskip


\end{document}